\documentclass[runningheads]{llncs}
\usepackage{graphicx}
\usepackage{url}
\usepackage{amsmath,amssymb,amsfonts}
\usepackage{booktabs}
\usepackage{xcolor}
\usepackage{hyperref}
\usepackage{multirow}
\usepackage{bm}
\usepackage[vlined,ruled]{algorithm2e}

\DeclareMathOperator*{\argmin}{arg\,min}

\SetKwRepeat{Do}{do}{while}

\begin{document}
\title{Enhancing Hyper-To-Real Space Projections Through Euclidean Norm Meta-Heuristic Optimization\thanks{The authors would like to thank S\~ao Paulo Research Foundation (FAPESP) grants \#2013/07375-0, \#2014/12236-1, \#2017/25908-6, \#2019/02205-5, \#2019/07825-1, and \#2019/07665-4, and National Council for Scientific and Technological Development (CNPq) grants \#307066/2017-7 and \#427968/2018-6.}}
\titlerunning{Enhancing Space Projections Through Meta-Heuristic Optimization}
%
\author{Luiz C. F. Ribeiro\orcidID{0000-0003-1265-0273} \and Mateus Roder\orcidID{0000-0002-3112-5290} \and Gustavo H. de Rosa\orcidID{0000-0002-6442-8343} \and Leandro A. Passos\orcidID{0000-0003-3529-3109} \and Jo\~ao P. Papa\orcidID{0000-0002-6494-7514}}
\authorrunning{Luiz C. F. Ribeiro et al.}
%
\institute{Department of Computing, S\~ao Paulo State University, Bauru, Brazil \\
\email{\{luiz.felix, mateus.roder, gustavo.rosa, leandro.passos, joao.papa\}@unesp.br}}
\maketitle              
\begin{abstract}
\begin{sloppypar}
The continuous computational power growth in the last decades has made solving several optimization problems significant to humankind a tractable task; however, tackling some of them remains a challenge due to the overwhelming amount of candidate solutions to be evaluated, even by using sophisticated algorithms. In such a context, a set of nature-inspired stochastic methods, called meta-heuristic optimization, can provide robust approximate solutions to different kinds of problems with a small computational burden, such as derivative-free real function optimization. Nevertheless, these methods may converge to inadequate solutions if the function landscape is too harsh, e.g., enclosing too many local optima. Previous works addressed this issue by employing a hypercomplex representation of the search space, like quaternions, where the landscape becomes smoother and supposedly easier to optimize. Under this approach, meta-heuristic computations happen in the hypercomplex space, whereas variables are mapped back to the real domain before function evaluation. Despite this latter operation being performed by the Euclidean norm, we have found that after the optimization procedure has finished, it is usually possible to obtain even better solutions by employing the Minkowski $p$-norm instead and fine-tuning $p$ through an auxiliary sub-problem with neglecting additional cost and no hyperparameters. Such behavior was observed in eight well-established benchmarking functions, thus fostering a new research direction for hypercomplex meta-heuristic optimization.
\end{sloppypar}

\end{abstract}
\section{Introduction}
\label{s.introduction}

Humanity sharpened their mathematical skills over several years of evolution by researching and studying formal and elegant tools to model world events' behavior. In such a context, when dealing with non-trivial problems, it is common to apply mathematical programming to overcome the before-mentioned tasks or even to streamline the process. Furthermore, once any prior knowledge might not be available, mathematical programming, commonly known as optimization~\cite{Torn:89}, provides an attractive approach to tackle the burden of empirical setups.


In the past decades, a new optimization paradigm called meta-heuristic has been used to solve several optimization problems~\cite{Yang:11}. Essentially, a meta-heuristic is a high-level abstraction of a procedure that generates and selects a heuristic that aims to provide a feasible solution to the problem. It combines concepts of \emph{exploration} and \emph{exploitation}, i.e., globally searching throughout the space and enhancing a promising solution based on its neighbors, respectively, constituted of complex learning procedures and simple searches usually inspired by biological behaviors. Additionally, they do not require specific domain knowledge and provide mechanisms to avoid susceptibility to local optima convergence. 

Although meta-heuristic techniques seem to be an exciting proposal, they still might perform poorly on challenging objective functions, being trapped in local optima and not achieving the most suitable solutions. Some attempts as hybrid variants~\cite{Li:20}, aging mechanisms~\cite{Deng:21}, and fitness landscape analysis~\cite{Engelbrecht:21} try to deal with this issue. Relying on more robust search spaces, such as representing each decision variable as a hypercomplex number, is an alternative approach that is not fully explored in the literature.

One can perceive that handling hypercomplex spaces is based on the likelihood of having more natural fitness landscapes, although mathematically not proved yet. The most common representations are the quaternions~\cite{Hamilton:44} and octonions~\cite{Graves:45}, which have compelling traits to describe the object's orientation in $n$-dimensional search spaces, being extremely useful in performing rotations in such spaces~\cite{Hart:94}. These representations have been successfully used in different areas, as in deep learning~\cite{PapaASC:17}, feature selection~\cite{Rosa:20}, special relativity~\cite{DeLeo:96} and quantum mechanics~\cite{Finkelstein:62}. Regarding meta-heuristic optimization, interesting results have been achieved in global optimization~\cite{FisterESA:13,PapaANNPR:16,Passos:19}, although not yet mathematically guaranteed.

Notwithstanding, hypercomplex optimization also has its particular problems, i.e., before attempting to feed quaternions or octonions to a real-valued objective function, one needs to project their values onto a real-valued space, usually accomplished by the Euclidean norm function. However, to the best of our knowledge, there is no work in the literature regarding how using the standard Euclidean norm function might affect the loss of information when projecting one space onto another. Thus, we are incredibly interested in exploring the possibility of employing the $p$-norm function and finding the most suitable $p$ value that minimizes the loss of information throughout the projection.

In this work, we investigate how employing the $p$-norm to refine the solution found by a standard hypercomplex meta-heuristic can affect the obtained result. In short, we optimize a real function using the standard quaternion-based variant of the Particle Swarm Algorithm (Q-PSO)~\cite{papaNIAAO:18}, i.e., meta-heuristic operations are performed in the hypercomplex space. In contrast, decision variables are mapped to the real domain through the Euclidean Norm for function evaluation. Notwithstanding, the best solution is refined by finding a more suitable projection between domains using the $p$-norm. The rationale for this decision lies in the fact that this operation is a Euclidean norm generalization. Hence we resort to fine-tuning new, yet not explored, hyperparameter in the optimization procedure, thus allowing more robust solutions to be found. Regardless, such a procedure can be applied to any hypercomplex-based meta-heuristic. Therefore, this work's main contributions are twofold: (i) to introduce a generic and inexpensive procedure to refine solutions found by hypercomplex meta-heuristics; and (ii) to foster research regarding how to map better hypercomplex to real values in the context of meta-heuristic optimization.

The remainder of this paper is organized as follows. Sections~\ref{s.hypercomplex} and~\ref{s.opt} present the theoretical background related to hypercomplex-based spaces (quaternions and Minkowski $p$-norm) and meta-heuristic optimization, respectively. Section~\ref{s.methodology} discusses the methodology adopted in this work, while Section~\ref{s.experiments} presents the experimental results. Finally, Section~\ref{s.conclusion} states conclusions and future works\footnote{The source code is available online at \url{https://github.com/lzfelix/lio}.}.

\section{Hypercomplex Representation}
\label{s.hypercomplex}

A quaternion $q$ is a hypercomplex number, composed of real and complex parts, being $q=a+bi+cj+dk$, where $a,b,c,d\in\mathbb{R}$ and $i,j,k$ are fundamental quaternions units. The basis equation that defines what a quaternion looks like is described as follows:

\begin{equation}
    i^2=j^2=k^2=ijk=-1.
\end{equation}

Essentially, a quaternion $q$ is a four-dimensional space representation over the real numbers, i.e., $\mathbb{R}^4$. Given two arbitrary quaternions $q_1=a+bi+cj+dk$ and $q_2=\alpha+\beta i+\gamma j+\delta k$ and a scalar $\kappa \in \mathbb{R}$, we define the quaternion algebra~\cite{Eberly:02} used throughout this work:


\begin{equation}
\begin{split}
    q_1+q_2 & =(a+bi+cj+dk)+(\alpha+\beta i+\gamma j+\delta k) \\
            &=(a+\alpha)+(b+\beta)i+(c+\gamma)j+(d+\delta)k,
    \label{e.q_add}
\end{split}
\end{equation}

\begin{equation}
\begin{split}
    q_1-q_2 & =(a+bi+cj+dk)-(\alpha+\beta i+\gamma j+\delta k) \\
            &=(a-\alpha)+(b-\beta)i+(c-\gamma)j+(d-\delta)k,
    \label{e.q_sub}
\end{split}
\end{equation}

\begin{equation}
\begin{split}
    \kappa q_1  &= \kappa (a+bi+cj+dk) \\
                &= \kappa a + (\kappa b)i + (\kappa c)j + (\kappa d)k.
    \label{e.q_scale}
\end{split}    
\end{equation}

\subsection{Minkowski $p$-norm}
\label{ss.mpNorm}

Another crucial operator that needs to be defined is the $p$-norm, which is responsible for mapping hypercomplex values to real numbers. Let $q$ be a hypercomplex number with real coefficients $\left\{z_d\right\}_{d=0}^{D-1}$, one can compute the Minkowski $p$-norm as follows:

\begin{equation}
\label{e.norm}
    \|q\|_p = \left( \sum_{d=0}^{D-1} |z_{d}|^p \right)^{1/p},
\end{equation}
where $D$ is the number of dimensions of the space ($2$ for complex numbers, and $4$ for quaternions, for instance) and $p \geq 1$. Common values for the latter variable are $1$ or $2$ for the Taxicab and Euclidean norms, respectively. Hence, one can see the $p$-norm as a generalization of such norm  operators.

\section{Meta-Heuristic Optimization}
\label{s.opt}

Optimization is the task of selecting a solution that best fits a function among a set of possible solutions. Several methods have been applied in this context, such as grid-search and gradient-based methods. Nevertheless, these methods carry a massive amount of computation, leading to burdened states in more complex problems, e.g., exponential and NP-complete problems.

An attempt to overcome such behaviors is to employ a meta-heuristic-based approach. Meta-heuristic techniques are nature-inspired stochastic algorithms that mimic an intelligence behavior, often observed in groups of animals, humans, or nature. Such approaches combine exploration and exploitation mechanisms in order to achieve sub-optimal solutions with low effort.

In this work, we employed the quaternion variant of the state-of-the-art Particle Swarm Optimization (PSO)~\cite{Kennedy:11} algorithm for function optimization. On the other hand, since fine-tuning the $p$ hyperparameter is a single-variable optimization task with a small search interval, we resort to the hyperparameter-less Black Hole (BH)~\cite{Hatamlou:13} algorithm.

\section{Methodology}
\label{s.methodology}

This section discusses how the presented meta-heuristics can be combined with quaternions to perform the so-called ``hypercomplex-based meta-heuristic optimization''. The proposed approach designated ``Last Iteration Optimization'' (LIO) is presented along with the considered benchmarking functions to evaluate it and the experimental setup.

\subsection{Hypercomplex Optimization}
\label{ss.hypercomplex-opt}

In their original formulation, meta-heuristic algorithms were conceived to optimize real-valued target functions with multiple real parameters. However, one may decide to represent each decision variable as quaternions.

In this case, each decision variable is represented by a quaternion with its real coefficients randomly initialized from a uniform distribution in the interval $[0, 1]$. Furthermore, the mapping from quaternions to real numbers for function evaluation becomes a paramount operation, which is usually carried out through the Euclidean norm. Still, care must be taken to ensure that this transformation does not yield numbers outside the feasibility region. Hence, hypercomplex coefficients are clipped individually to the real interval $[0, 1]$ and the mapping for each decision variable is performed by the following mapping function:

\begin{equation}
\begin{split}
    \bm{\hat{q}}_{j} &= M(\bm{q}_j, p) \\ &= \bm{l}_j + \left( \bm{u}_j - \bm{l}_j \right) \frac{\|\bm{q}_{j}\|_p}{D^{1/p}},
\label{e.mapping}
\end{split}
\end{equation}
such that ${j = \{1, 2, \ldots, n\}}$, $D$ is the number of hypercomplex dimensions (4 for quaternions), $\bm{l}_j$ and $\bm{u}_j$ are the lower and upper bounds for each decision variable, respectively, and $p=2$ in this particular case.

\subsection{Last Iteration Optimization}
\label{ss.proposed}

\begin{sloppypar}
The main goal of this work consists of refining the solution found by a hypercomplex-based meta-heuristic using a low-cost procedure. To such an extent, given a fitness function $f: \mathbb{R}^{n} \rightarrow \mathbb{R}$, we first optimize it through the Q-PSO algorithm, which consists in representing each decision variable as a quaternion with the relations defined in Equations~\ref{e.q_add}, \ref{e.q_sub}, and~\ref{e.q_scale}. Once this step is finished, we have the best candidate solution $\bm{q}^\star$ with a real representation $\bm{\hat{q}}^\star \in \mathbb{R}^n$, which is obtained through Equation~\ref{e.mapping} with $p=2$. Shortly, one can compute the best solution fitness $\mu$ as follows:
\end{sloppypar}

\begin{equation}
    \mu = f\Big( M(\bm{q}^\star_1, 2), M(\bm{q}^\star_2, 2), \ldots, M(\bm{q}^\star_n, 2) \Big).
\end{equation}
where $M(\cdot)$ is computed according to Equation~\ref{e.mapping}.

We propose a second phase to the optimization pipeline, where the best solution found is $\bm{q}^\star$ is kept fixed, while the hyper-parameter $p$ is unfrozen. Such an approach allows obtaining a better real representation of $\bm{q}^\star$, which translates to an even smaller fitness value $\mu^{\star}$. Namely, we aim at solving the following auxiliary optimization problem:

\begin{equation}
\begin{aligned}
    p^\star =& \argmin_p \, f \Big(M(\bm{q}^\star_1, p), M(\bm{q}^\star_2, p), \ldots, M(\bm{q}^\star_n, p) \Big), \\
      & \text{st. } 1 \leq p \leq p_{\max},
      \label{e.p_optimization}
\end{aligned}
\end{equation}
where $p_{\max}$ denotes the maximum possible value for parameter $p$. If $p_{\max} = 2$, for instance, the problem consists in finding a suitable norm between the Taxicab and Euclidean ones.

Since the new search interval is usually small, as it is going to be discussed in Section~\ref{ss.setup}, we resort to the traditional BH algorithm since it does not contain hyperparameters to be tuned, thus making the process even simpler. As this procedure is performed for a single decision variable in a small search space, the time spent in this phase is negligible compared to the Q-PSO step. Furthermore, since this new step is performed as the new last iteration of the optimization pipeline, we name it Last Iteration Optimization (LIO).

\subsection{Benchmarking Functions}
\label{ss.benchmark}

Table~\ref{t.bench_functions} introduces the eight benchmarking functions used to evaluate the proposed approach.

\begin{table}[!ht]
	\centering
	\caption{Benchmarking functions.}
	\label{t.bench_functions}
	\renewcommand{\arraystretch}{1.35}
    \resizebox{\textwidth}{!}{
	\begin{tabular}{lccc}
		\toprule
		\textbf{Function} & \textbf{Equation} & \textbf{Bounds} & $\mathbf{f(\bm{x}^\ast)}$ \\
		\midrule
		Sphere & $f_1(x) = \sum\limits_{i=1}^n x_i^2 $ & $-10 \leq x_i \leq 10$ & 0\\\\

		Csendes & $f_2(x) = \sum_{i=1}^n x_i^6\left(2+sin\frac{1}{x_i}\right)$ & $-1 \leq x_i \leq 1$ & 0\\\\

		Salomon & $f_3(x) = 1 - \cos(2\pi\sqrt{\sum_{i=1}^{n}x^2_i})+0.1\sqrt{\sum_{i=1}^{n}x^2_i}$ & $-100 \leq x_i \leq 100$ & 0\\\\

		Ackley \#1 & $f_4(x) = -20e^{-0.02\sqrt{n^{-1}\sum_{i=1}^n x_i^2}}-e^{n^{-1} \sum_{i=1}^n cos(2\pi x_i)}+20+e$ & $-35 \leq x_i \leq 35$ & 0\\\\

		Alpine \#1 & $f_5(x) = \sum_{i=1}^{n}|x_i sin(x_i)+0.1x_i|$ & $-10 \leq x_i \leq 10$ & 0\\\\

		Rastrigin  & $f_6(x) = 10n + \sum_{i=1}^n\left[x_i^2-10cos(2\pi x_i)\right]$ & $-5.12 \leq x_i \leq 5.12$ & 0\\\\

		Schwefel & $f_7(x) =\left(\sum\limits_{i=1}^n x_i^2\right)^{\sqrt{\pi}}$ & $-100 \leq x_i \leq 100$ & 0\\\\

		Brown  & $f_8(x) = \sum_{i=1}^{n-1} \left[ (x_i^2)^{(x_{i+1}^2+1)}+(x_{i+1}^2)^{(x_i^2+1)} \right]$ & $-1 \leq x_i \leq 4$ & 0\\
		\bottomrule
	\end{tabular}}
\end{table}

\subsection{Experimental Setup}
\label{ss.setup}

The proposed approach divides function optimization into two parts: global and fine-tuning phases, which correspond to finding $\mu$ using Q-PSO and $\mu^\star$ by solving Equation~\ref{e.p_optimization} through the BH algorithm. 

Regarding the first phase, we use the same experimental setup from~\cite{Passos:19}. Namely, each benchmark function is optimized with $n \in \{10, 25, 50, 100\}$ decision variables, for $(2000 \cdot n)$ iterations using $100$ agents. As the amount of iterations grows considerably fast, we adapt to the early stopping mechanism. Such a strategy allows detecting if the optimization is stuck for too long in a local optimum and unlikely to find a better solution, saving computational time. If the difference of fitness between two consecutive iterations is smaller than $\delta = 10^{-5}$ for $50$ iterations or more, the optimization is halted, and the best fitness found so far is deemed the solution. Despite these values being determined empirically, they often present the same results as those obtained using all available iterations, despite using, at most $4\%$ of all iterations for the extreme case when $n=100$. For the Q-PSO hyperparameters we use $w=0.7$, $c_1 = c_2 = 1.7$, as well established in the literature.

In the second phase, optimization is performed with ${p_{\max}=5}$, using $20$ agents for $50$ iterations, which were determined on preliminary experiments. Further, we do not rely on early stopping for this phase since it is performed much faster than the previous one. Finally, we compare the results obtained by Q-PSO and Q-PSO with LIO. Each experiment is executed 15 times, and the best results with significance smaller than $0.05$, according to the Wilcoxon signed-rank~\cite{Wilcoxon:45}, are highlighted in bold. Regarding the implementation, we used Opytimizer~\cite{rosa2019opytimizer} library.

\section{Experimental Results}
\label{s.experiments}

Experimental results are presented in Table~\ref{t.results}, where the average fitness values obtained by Q-PSO are compared against their refined versions, computed with LIO. More specifically, we ran Q-PSO, stored the results, and continued the LIO (denoted as Q-PSO + LIO).

\begin{table}[!ht]
    \centering
    \caption{Best fitness found by varying the number of decision variables for each benchmark function.}
    \label{t.results}
    \renewcommand{\arraystretch}{1.675}
    \resizebox{\textwidth}{!}{
    \begin{tabular}{lcllcrr}
    \toprule
    \multicolumn{1}{c}{\textbf{Functions}} &\multicolumn{1}{c}{\textbf{Dimensions}} &\multicolumn{1}{c}{\textbf{Q-PSO}} &\multicolumn{1}{c}{\textbf{Q-PSO + LIO}} &\multicolumn{1}{c}{$\bm{p}$} &\multicolumn{1}{c}{\textbf{Q-PSO time (s)}} &\multicolumn{1}{c}{\textbf{LIO time (s)}}
    \\
    \midrule
    \multirow{4}{*}{Sphere} & 
    10 	 & 
        $1.3447 \cdot 10^{-7} \pm 1.8964 \cdot 10^{-7}$	 & 
        $\bm{1.2169 \cdot 10^{-7} \pm 1.6903 \cdot 10^{-7}}$	 & 
        1.99 $\pm$ 5.78	 & 
        5.66 $\pm$ 0.35	&
        0.29 $\pm$ 0.01	\\
    & 25 	 & 
        $3.1993 \cdot 10^{-1} \pm 1.9855 \cdot 10^{-1}$	 & 
        $\bm{3.0657 \cdot 10^{-1} \pm 1.9018 \cdot 10^{-1}}$	 & 
        1.98 $\pm$ 0.06	 & 
        33.08 $\pm$ 3.93	&
        0.30 $\pm$ 0.01	\\
    & 50 	 & 
        $5.3962 \cdot 10^{0} \pm 2.0896 \cdot 10^{0}$	 & 
        $\bm{5.3205 \cdot 10^{0} \pm 2.0266 \cdot 10^{0}}$	 & 
        2.01 $\pm$ 0.19	 & 
        77.55 $\pm$ 14.45	&
        0.37 $\pm$ 0.01	\\
    & 100	 & 
        $2.3679 \cdot 10^{1} \pm 3.3931 \cdot 10^{0}$	 & 
        $\bm{2.3433 \cdot 10^{1} \pm 3.4593 \cdot 10^{0}}$	 & 
        1.84 $\pm$ 0.38	 & 
        158.04 $\pm$ 24.79	&
        0.40 $\pm$ 0.01	\\
    \midrule
    \multirow{4}{*}{Csendes} & 
    10 	 & 
        $4.1345 \cdot 10^{-13} \pm 1.2771 \cdot 10^{-12}$	 & 
        $\bm{2.3502 \cdot 10^{-13} \pm 6.5480 \cdot 10^{-13}}$	 & 
        1.99 $\pm$ 0.00	 & 
        2.43 $\pm$ 0.12	&
        0.31 $\pm$ 0.01	\\
    & 25 	 & 
        $6.2160 \cdot 10^{-7} \pm 6.2587 \cdot 10^{-7}$	 & 
        $\bm{5.8670 \cdot 10^{-7} \pm 5.6680 \cdot 10^{-7}}$	 & 
        1.97 $\pm$ 0.11	 & 
        6.16 $\pm$ 0.35	&
        0.34 $\pm$ 0.02	\\
    & 50 	 & 
        $8.5587 \cdot 10^{-6} \pm 6.5273 \cdot 10^{-6}$	 & 
        $\bm{8.3406 \cdot 10^{-6} \pm 6.2563 \cdot 10^{-6}}$	 & 
        2.01 $\pm$ 0.05	 & 
        13.31 $\pm$ 0.71	&
        0.39 $\pm$ 0.01	\\
    & 100	 & 
        $4.9290 \cdot 10^{-5} \pm 2.3214 \cdot 10^{-5}$	 & 
        $\bm{4.8460 \cdot 10^{-5} \pm 2.3491 \cdot 10^{-5}}$	 & 
        1.94 $\pm$ 0.15	 & 
        31.88 $\pm$ 2.95	&
        0.48 $\pm$ 0.02	\\
    \midrule
    \multirow{4}{*}{Salomon} & 
    10 	 & 
        $\bm{3.4654 \cdot 10^{-1} \pm 1.0242 \cdot 10^{-1}}$	 & 
        $3.4655 \cdot 10^{-1} \pm 1.0243 \cdot 10^{-1}$	 & 
        2.04 $\pm$ 0.08	 & 
        4.97 $\pm$ 0.58	&
        0.30 $\pm$ 0.02	\\
    & 25 	 & 
        $2.0332 \cdot 10^{0} \pm 4.3919 \cdot 10^{-1}$	 & 
        $\bm{2.0199 \cdot 10^{0} \pm 4.4000 \cdot 10^{-1}}$	 & 
        2.07 $\pm$ 0.29	 & 
        11.73 $\pm$ 1.78	&
        0.31 $\pm$ 0.01	\\
    & 50 	 & 
        $\bm{4.1332 \cdot 10^{0} \pm 5.6529 \cdot 10^{-1}}$	 & 
        $\bm{4.1000 \cdot 10^{0} \pm 5.4903 \cdot 10^{-1}}$	 & 
        2.22 $\pm$ 0.52	 & 
        24.11 $\pm$ 3.30	&
        0.36 $\pm$ 0.01	\\
    & 100	 & 
        $6.3799 \cdot 10^{0} \pm 7.2682 \cdot 10^{-1}$	 & 
        $\bm{6.3665 \cdot 10^{0} \pm 7.0016 \cdot 10^{-1}}$	 & 
        2.07 $\pm$ 0.46	 & 
        58.78 $\pm$ 9.67	&
        0.44 $\pm$ 0.02	\\
    \midrule
    \multirow{4}{*}{Ackley \#1} & 
    10 	 & 
        $\bm{8.7839 \cdot 10^{-1} \pm 2.9911 \cdot 10^{-1}}$	 & 
        $8.7843 \cdot 10^{-1} \pm 2.9914 \cdot 10^{-1}$	 & 
        2.00 $\pm$ 0.00	 & 
        7.61 $\pm$ 1.23	&
        0.35 $\pm$ 0.02	\\
    & 25 	 & 
        $1.2330 \cdot 10^{0} \pm 2.5165 \cdot 10^{-1}$	 & 
        $\bm{1.2293 \cdot 10^{0} \pm 2.5283 \cdot 10^{-1}}$	 & 
        1.99 $\pm$ 0.00	 & 
        35.09 $\pm$ 5.11	&
        0.38 $\pm$ 0.02	\\
    & 50 	 & 
        $1.8135 \cdot 10^{0} \pm 1.8909 \cdot 10^{-1}$	 & 
        $\bm{1.8062 \cdot 10^{0} \pm 1.9024 \cdot 10^{-1}}$	 & 
        2.00 $\pm$ 0.01	 & 
        61.51 $\pm$ 9.02	&
        0.40 $\pm$ 0.02	\\
    & 100	 & 
        $2.2038 \cdot 10^{0} \pm 1.0338 \cdot 10^{-1}$	 & 
        $\bm{2.2006 \cdot 10^{0} \pm 1.0280 \cdot 10^{-1}}$	 & 
        2.00 $\pm$ 0.01	 & 
        125.12 $\pm$ 14.76	&
        0.49 $\pm$ 0.03	\\
    \midrule
    \multirow{4}{*}{Alpine \#1} & 
    10 	 & 
        $\bm{7.9560 \cdot 10^{-2} \pm 1.2551 \cdot 10^{-1}}$	 & 
        $\bm{7.9515 \cdot 10^{-2} \pm 1.2565 \cdot 10^{-1}}$	 & 
        2.00 $\pm$ 0.00	 & 
        9.95 $\pm$ 2.36	&
        0.29 $\pm$ 0.02	\\
    & 25 	 & 
        $\bm{2.2345 \cdot 10^{0} \pm 1.3898 \cdot 10^{0}}$	 & 
        $\bm{2.2240 \cdot 10^{0} \pm 1.3957 \cdot 10^{0}}$	 & 
        2.01 $\pm$ 0.05	 & 
        27.84 $\pm$ 3.47	&
        0.31 $\pm$ 0.02	\\
    & 50 	 & 
        $\bm{1.2124 \cdot 10^{1} \pm 4.1844 \cdot 10^{0}}$	 & 
        $\bm{1.2088 \cdot 10^{1} \pm 4.1675 \cdot 10^{0}}$	 & 
        2.00 $\pm$ 0.05	 & 
        63.12 $\pm$ 11.53	&
        0.35 $\pm$ 0.01	\\
    & 100	 & 
        $\bm{2.7375 \cdot 10^{1} \pm 7.7322 \cdot 10^{0}}$	 & 
        $\bm{2.7322 \cdot 10^{1} \pm 7.7351 \cdot 10^{0}}$	 & 
        1.95 $\pm$ 0.20	 & 
        124.72 $\pm$ 20.72	&
        0.42 $\pm$ 0.02	\\
    \midrule
    \multirow{4}{*}{Rastrigin} & 
    10 	 & 
        $\bm{1.1608 \cdot 10^{1} \pm 5.1079 \cdot 10^{0}}$	 & 
        $\bm{1.1608 \cdot 10^{1} \pm 5.1079 \cdot 10^{0}}$	 & 
        2.00 $\pm$ 0.00	 & 
        7.21 $\pm$ 0.60	&
        0.33 $\pm$ 0.02	\\
    & 25 	 & 
        $3.7845 \cdot 10^{1} \pm 1.3993 \cdot 10^{1}$	 & 
        $\bm{3.7673 \cdot 10^{1} \pm 1.4040 \cdot 10^{1}}$	 & 
        1.99 $\pm$ 0.01	 & 
        39.67 $\pm$ 5.66	&
        0.33 $\pm$ 0.02	\\
    & 50 	 & 
        $1.4677 \cdot 10^{2} \pm 2.3428 \cdot 10^{1}$	 & 
        $\bm{1.4506 \cdot 10^{2} \pm 2.3524 \cdot 10^{1}}$	 & 
        2.00 $\pm$ 0.04	 & 
        94.89 $\pm$ 20.97	&
        0.37 $\pm$ 0.02	\\
    & 100	 & 
        $4.8812 \cdot 10^{2} \pm 4.7400 \cdot 10^{1}$	 & 
        $\bm{4.8437 \cdot 10^{2} \pm 4.7186 \cdot 10^{1}}$	 & 
        1.99 $\pm$ 0.05	 & 
        208.48 $\pm$ 39.45	&
        0.44 $\pm$ 0.02	\\
    \midrule
    \multirow{4}{*}{Schwefel} & 
    10 	 & 
        $4.9247 \cdot 10^{-9} \pm 9.7025 \cdot 10^{-9}$	 & 
        $\bm{3.5510 \cdot 10^{-9} \pm 7.0349 \cdot 10^{-9}}$	 & 
        1.99 $\pm$ 7.36	 & 
        6.20 $\pm$ 0.43	&
        0.31 $\pm$ 0.01	\\
    & 25 	 & 
        $1.5213 \cdot 10^{3} \pm 2.5509 \cdot 10^{3}$	 & 
        $\bm{1.1048 \cdot 10^{3} \pm 1.2858 \cdot 10^{3}}$	 & 
        1.97 $\pm$ 0.07	 & 
        64.56 $\pm$ 12.69	&
        0.31 $\pm$ 0.01	\\
    & 50 	 & 
        $9.4460 \cdot 10^{4} \pm 5.9571 \cdot 10^{4}$	 & 
        $\bm{8.8915 \cdot 10^{4} \pm 5.3935 \cdot 10^{4}}$	 & 
        1.88 $\pm$ 0.23	 & 
        163.12 $\pm$ 52.00	&
        0.35 $\pm$ 0.01	\\
    & 100	 & 
        $9.4876 \cdot 10^{5} \pm 3.8055 \cdot 10^{5}$	 & 
        $\bm{9.0762 \cdot 10^{5} \pm 4.0064 \cdot 10^{5}}$	 & 
        2.29 $\pm$ 0.56	 & 
        349.40 $\pm$ 112.34	&
        0.40 $\pm$ 0.02	\\
    \midrule
    \multirow{4}{*}{Brown} & 
    10 	 & 
        $3.3230 \cdot 10^{0} \pm 4.3978 \cdot 10^{0}$	 & 
        $\bm{1.0051 \cdot 10^{0} \pm 9.4007 \cdot 10^{-1}}$	 & 
        1.58 $\pm$ 0.37	 & 
        8.21 $\pm$ 3.22	&
        0.33 $\pm$ 0.01	\\
    & 25 	 & 
        $1.4622 \cdot 10^{1} \pm 7.9266 \cdot 10^{0}$	 & 
        $\bm{5.6664 \cdot 10^{0} \pm 2.9834 \cdot 10^{0}}$	 & 
        1.13 $\pm$ 0.11	 & 
        56.16 $\pm$ 13.62	&
        0.34 $\pm$ 0.02	\\
    & 50 	 & 
        $2.8192 \cdot 10^{2} \pm 1.0655 \cdot 10^{2}$	 & 
        $\bm{1.3212 \cdot 10^{2} \pm 5.4533 \cdot 10^{1}}$	 & 
        1.01 $\pm$ 0.02	 & 
        120.97 $\pm$ 36.77	&
        0.41 $\pm$ 0.03	\\
    & 100	 & 
        $1.8173 \cdot 10^{3} \pm 3.1725 \cdot 10^{2}$	 & 
        $\bm{1.2501 \cdot 10^{3} \pm 3.0196 \cdot 10^{2}}$	 & 
        1.01 $\pm$ 0.01	 & 
        245.45 $\pm$ 82.48	&
        0.49 $\pm$ 0.02	\\
    \bottomrule
    \end{tabular}}
\end{table}

\subsection{Overall Discussion}
\label{ss.exp_overall}

Experimental results provided in Table~\ref{t.results} confirm the robustness of the proposed approach since the Q-PSO + LIO outperformed the standard Q-PSO in the massive majority of benchmarking functions and configurations. One can highlight, for instance, that LIO obtained the best results overall, considering all dimensional configurations, in half of the functions, i.e., Sphere, Csendes, Schwefel, and Brown. Besides, Alpine1 and Rastrigin can also be deliberated, although Q-PSO obtained similar statistical results. Further, LIO also obtained the best results considering all functions over three-out-of-four configurations, i.e., $25$, $50$, and $100$ dimensions, being Q-PSO statistical similar in only two of them.

On the other hand, Q-PSO obtained the best results over two functions, i.e., Salomon and Rastrigin, considering a $10$-dimensional configuration. Such behavior is very interesting since Q-PSO performed better over two functions who share similar characteristics: both are continuous, differentiable, non-separable, scalable, and multimodal, contemplating the same dimensionality, which may denote some specific constraint to the model. 

Finally, as an overview, the proposed approach can significantly improve Q-PSO, with an almost insignificant computational burden, and whose growth is barely insignificant compared to the increase in the number of dimensions, as discussed in the next section.

\subsection{Computational Burden}
\label{ss.exp_burden}

Germane to this aspect, the results in Table~\ref{t.results} show that LIO takes significantly less time than the main meta-heuristic to be evaluated. This phenomenon is expected since the latter involves solving an optimization problem with a single real variable in a small search interval. Nonetheless, despite this simplicity, our results show promising results by performing such a task. In the worst-case scenario, i.e.,  Csendes function with 10 variables, LIO takes only $12.6\%$ of the time consumed by Q-PSO, which amounts to $0.31$ seconds, while decreasing the fitness value by a factor of $1.7$.

\subsection{How does $p$ Influence Projections?}
\label{ss.influence}

From the results in Table~\ref{t.results}, one can highlight the variation in $p$-norm value. As expected, such a variable is highly correlated to the optimization performance, since small changes in its value resulted in better functions minima. On the other hand, one can notice that expressive changes in $p$ may also support performance improvement, as in Brown function. Besides that, as $p$ is changed, the mapping process, i.e., the projection, from the hypercomplex space to the real one becomes ``less aggressive'' to the latter, since the proposed approach gives margin to a smooth fit for the values obtained in the former space.

Therefore, examining the performance on the optimization functions, one can observe that employing LIO's projection, different optimization landscapes are achieved, and such a process provides better value's representation from the hypercomplex search space. It is worth observing that for Rastrigin, Alpine \#1, and Ackley \#1 functions, LIO found optimal $p$ values with mean $2$ and minimal standard deviations, thus showing this parameter's sensitiveness for some benchmarking functions. Moreover, only LIO optimization for the Schwefel function with 10 dimensions showed a large standard deviation for this hyperparameter. In contrast, in the remaining cases, there was no norm larger than 3, suggesting that in further experiments, and even smaller search intervals (with $p_{\max}=3$, for instance) could be employed.


\section{Conclusion}
\label{s.conclusion}

In this work, we introduced the Last Iteration Optimization (LIO) procedure, which consists of refining the solution found by a hypercomplex-based meta-heuristic optimization algorithm by solving a low-cost hyperparameter-less auxiliary problem after the primary heuristic has found the best candidate solution. Such a procedure provided robust results in various benchmarking functions, showing statistically significant gains in 24 out of 32 experiments, over functions with diverse characteristics. Since LIO has a low computational burden and is easy to implement, it can be readily incorporated into other works.

In future studies, we intend to investigate how changing the $p$ parameter during the global optimization procedure can affect the obtained results. Furthermore, LIO can be extended to find a different $p$ for each decision variable, making it more flexible, and even other functions can be employed (or learned) to perform the hypercomplex-to-real mapping process. Ultimately, fine-tuning the $p$ hyper-parameter of the Minkowski norm opens new research directions for hypercomplex-based meta-heuristic function optimization methods.
%
%
%
\bibliographystyle{splncs04}
\bibliography{references}

\begin{thebibliography}{10}
\providecommand{\url}[1]{\texttt{#1}}
\providecommand{\urlprefix}{URL }
\providecommand{\doi}[1]{https://doi.org/#1}

\bibitem{Deng:21}
Deng, L., Sun, H., Li, C.: Jdf-de: A differential evolution with jrand number
  decreasing mechanism and feedback guide technique for global numerical
  optimization. Applied Intelligence  \textbf{51}(1),  359--376 (2021)

\bibitem{Eberly:02}
Eberly, D.: Quaternion algebra and calculus. Tech. rep., Magic Software (2002)

\bibitem{Engelbrecht:21}
Engelbrecht, A., Bosman, P., Malan, K.: The influence of fitness landscape
  characteristics on particle swarm optimisers. Natural Computing pp. 1--11
  (2021)

\bibitem{Finkelstein:62}
Finkelstein, D., Jauch, J.M., Schiminovich, S., Speiser, D.: Foundations of
  quaternion quantum mechanics. Journal of mathematical physics  \textbf{3}(2),
   207--220 (1962)

\bibitem{FisterESA:13}
Fister, I., Yang, X.S., Brest, J., Jr., I.F.: Modified firefly algorithm using
  quaternion representation. Expert Systems with Applications  \textbf{40}(18),
   7220--7230 (2013). \doi{http://dx.doi.org/10.1016/j.eswa.2013.06.070}

\bibitem{Graves:45}
Graves, J.T.: On a connection between the general theory of normal couples and
  the theory of complete quadratic functions of two variables. Philosophical
  Magazine  \textbf{26}(173),  315--320 (1845)

\bibitem{Hamilton:44}
Hamilton, W.R.: On quaternions; or on a new system of imaginaries in algebra.
  The London, Edinburgh, and Dublin Philosophical Magazine and Journal of
  Science  \textbf{25}(163),  10--13 (1844)

\bibitem{Hart:94}
Hart, J.C., Francis, G.K., Kauffman, L.H.: Visualizing quaternion rotation. ACM
  Transactions on Graphics (TOG)  \textbf{13}(3),  256--276 (1994)

\bibitem{Hatamlou:13}
Hatamlou, A.: Black hole: A new heuristic optimization approach for data
  clustering. Information sciences  \textbf{222},  175--184 (2013)

\bibitem{Kennedy:11}
Kennedy, J.: Particle swarm optimization. In: Encyclopedia of machine learning,
  pp. 760--766. Springer (2011)

\bibitem{DeLeo:96}
Leo, S.D.: Quaternions and special relativity. Journal of Mathematical Physics
  \textbf{37}(6),  2955--2968 (1996)

\bibitem{Li:20}
Li, J., Lei, H., Alavi, A.H., Wang, G.G.: Elephant herding optimization:
  variants, hybrids, and applications. Mathematics  \textbf{8}(9), ~1415 (2020)

\bibitem{PapaANNPR:16}
Papa, J.P., Pereira, D.R., Baldassin, A., Yang, X.S.: On the harmony search
  using quaternions. In: Schwenker, F., Abbas, H.M., El-Gayar, N., Trentin, E.
  (eds.) Artificial Neural Networks in Pattern Recognition: 7th IAPR TC3
  Workshop, ANNPR. pp. 126--137. Springer International Publishing, Cham (2016)

\bibitem{PapaASC:17}
Papa, J.P., Rosa, G.H., Pereira, D.R., Yang, X.S.: Quaternion-based deep belief
  networks fine-tuning. Applied Soft Computing  \textbf{60},  328--335 (2017)

\bibitem{papaNIAAO:18}
Papa, J.P., de~Rosa, G.H., Yang, X.S.: On the hypercomplex-based search spaces
  for optimization purposes. In: Nature-Inspired Algorithms and Applied
  Optimization, pp. 119--147. Springer (2018)

\bibitem{Passos:19}
Passos, L.A., Rodrigues, D., Papa, J.P.: Quaternion-based backtracking search
  optimization algorithm. In: 2019 IEEE Congress on Evolutionary Computation
  (CEC). pp. 3014--3021. IEEE (2019)

\bibitem{Rosa:20}
de~Rosa, G.H., Papa, J.P., Yang, X.S.: A nature-inspired feature selection
  approach based on hypercomplex information. Applied Soft Computing
  \textbf{94},  106453 (2020)

\bibitem{rosa2019opytimizer}
de~Rosa, G.H., Papa, J.P.: Opytimizer: A nature-inspired python optimizer
  (2019)

\bibitem{Torn:89}
T{\"o}rn, A., {\v{Z}}ilinskas, A.: Global optimization, vol.~350. Springer
  (1989)

\bibitem{Wilcoxon:45}
Wilcoxon, F.: Individual comparisons by ranking methods. Biometrics Bulletin
  \textbf{1}(6),  80--83 (1945)

\bibitem{Yang:11}
Yang, X.S.: Review of meta-heuristics and generalised evolutionary walk
  algorithm. International Journal of Bio-Inspired Computation  \textbf{3}(2),
  77--84 (2011)

\end{thebibliography}

\end{document}